\documentclass{article}
\usepackage[letterpaper, margin=1in]{geometry}
\usepackage[utf8]{inputenc}

\usepackage{color,soul}

\usepackage{natbib}
\usepackage{graphicx}
\usepackage{paralist}

\usepackage{amsmath}
\usepackage{amsthm}
\usepackage{amstext}
\usepackage{amsfonts}
\usepackage{mathrsfs}
\usepackage{amssymb}
\usepackage{nicefrac}
\usepackage{bm}
\usepackage{xfrac}
\usepackage{authblk}

\newtheorem{theorem}{Theorem}[section]

\theoremstyle{definition}
\newtheorem{definition}[theorem]{Definition}
\newtheorem{proposition}[theorem]{Proposition}
\newtheorem{example}[theorem]{Example}

\newtheorem{remark}[theorem]{Remark}

\newcommand{\fun}[2]{#1\!\left(#2\right)}

\newcommand{\PP}[2]{\fun{P_{#1\!}}{#2}} 

\newcommand{\given}{\vert} 
\newcommand{\transpose}[1]{#1^\mathsf{T}}

\newcommand{\size}[1]{|#1|} 
\newcommand{\Diag}[1]{\mathrm{Diag}\!\left(#1\right)}

\newcommand{\data}{D} 
\newcommand{\dataSpace}{\mathcal D} 
\newcommand{\dataSize}{\size{\dataSpace}}
\newcommand{\concept}{h} 
\newcommand{\conceptSpace}{\mathcal H}
\newcommand{\conceptSize}{\size{\conceptSpace}}
\newcommand{\TI}{\mathrm{TI}} 
\newcommand{\CI}{\mathrm{CI}} 

\newcommand{\permu}{\sigma}

\newcommand{\sumx}{\sum_{\data \in \dataSpace}}
\newcommand{\sumh}{\sum_{\concept \in \conceptSpace}}
\newcommand{\sumxi}{\sum_{i=1}^{\size{\dataSpace}}}
\newcommand{\sumhj}{\sum_{j=1}^{\size{\conceptSpace}}}

\newcommand{\LL}{\PP{L}{\concept\given\data}}
\newcommand{\LLpri}{\PP{L_0}{\concept}}
\newcommand{\LLmar}{\PP{L}{\data}}
\newcommand{\LLit}[1]{\PP{L_{#1}}{\concept\given\data}}
\newcommand{\LLitmar}[1]{\PP{L_{#1}}{\data}}
\newcommand{\TT}{\PP{T}{\data\given\concept}} 
\newcommand{\TTpri}{\PP{T_0}{\data}}
\newcommand{\TTmar}{\PP{T}{\concept}}
\newcommand{\TTit}[1]{\PP{T_{#1}}{\data\given\concept}}
\newcommand{\TTitmar}[1]{\PP{T_{#1}}{\concept}}
\newcommand{\Lik}{\PP{0}{\data\given\concept}}

\newcommand{\LLmat}{\mathbf{L}}
\newcommand{\LLmatit}[1]{\mathbf{L}^{(#1)}}
\newcommand{\LLvecpri}{\mathbf{a}}
\newcommand{\LLmatpri}{\Diag{\LLvecpri}}
\newcommand{\LLvecmar}{\mathbf{c}}
\newcommand{\LLvecmarit}[1]{\mathbf{c}^{(#1)}}
\newcommand{\TTmat}{\mathbf{T}} 
\newcommand{\TTmatit}[1]{\mathbf{T}^{(#1)}} 
\newcommand{\TTvecpri}{\mathbf{b}}

\newcommand{\TTmatpri}{\Diag{\TTvecpri}}
\newcommand{\TTvecmar}{\mathbf{d}}
\newcommand{\TTvecmarit}[1]{\mathbf{d}^{(#1)}}
\newcommand{\Likmat}{\mathbf{M}\,}

\newcommand{\Likmatconverge}{\mathbf{M}^{(\infty)}}

\title{Optimal Cooperative Inference}
\author{Scott Cheng-Hsin Yang \thanks{scott.cheng.hsin.yang@gmail.com}}
\author{Yue Yu}
\author{Arash Givchi}
\author{Pei Wang}
\author{Wai Keen Vong}
\author{Patrick Shafto \thanks{patrick.shafto@gmail.com}}
\affil{Department of Mathematics \& Computer Science, Rutgers University---Newark}
\date{}

\begin{document}

\maketitle

\begin{abstract}
Cooperative transmission of data fosters rapid accumulation of knowledge by efficiently combining experiences across learners. 
Although well studied in human learning and increasingly in machine learning, we lack formal frameworks through which we may reason about the benefits and limitations of cooperative inference. We present such a framework. 
We introduce novel indices for measuring the effectiveness of probabilistic 
and cooperative information transmission.
We relate our indices to the well-known Teaching Dimension in deterministic settings.
We prove conditions under which optimal cooperative inference can be achieved, including a representation theorem that constrains the form of inductive biases for learners optimized for cooperative inference.
We conclude by demonstrating how these principles may inform the design of machine learning algorithms and discuss implications for human and machine learning. 
\end{abstract}





\section{INTRODUCTION}



Learning through cooperation is a foundational principle underlying human-human, human-machine, and (potentially) machine-machine interaction. In human-human interaction, cooperative information sharing has long been viewed as a foundation to human language \citep{grice1975logic, goodman2013knowledge, Kao2014}, cognitive development \citep{csibra2009natural}, and cultural evolution \citep{tomasello1999,tomasello2005search}. Cooperative learning has appeared in human-machine interaction \citep{crandall2017}, social robotics \citep{thomaz2008teachable, knox2013training, chernova2014robot, thomaz2016computational, laskey2017comparing, bestick2016implicitly}, machine teaching \citep{zhu2013machine, Zhu2015, patil2014optimal, simard2017machine}, cooperative reinforcement learning \citep{hadfield2016cooperative, ho2016showing}, and deep neural networks \citep{lowe2017multi}. Despite the importance of cooperative selection of, and learning from, data, we are unaware of any theory of when or why cooperation may be effective for increasing learning and the transmission of knowledge.

In this paper we address this lack by introducing a measure of communication effectiveness in the cooperative setting.
The role that this measure plays in cooperative knowledge accumulation is analogous to the role that training and test errors play in traditional machine learning.
As training and test errors provide a framework for measuring how effectively a model selects the best model and generalizes, our new measure, Cooperative Index, provides a framework for measuring how effectively a model can be explained by way of examples from the data and for selecting models with inductive biases that are interpretable with respect to the data. 
We also use the measure to extend the Teaching Dimension \citep{goldman1995complexity, Zilles2008}---a classical measure of communication efficiency\footnote{Effectiveness is a measure of the quality of communication; efficiency is the size of the data necessary to reach a particularly effectiveness.}---from deterministic to probabilistic settings.
We show how analyzing this measure reveals the conditions, in terms of constraints on the learning model's inductive biases, under which cooperation may produce optimal communication. 

The paper is organized as follows: In Section~\ref{sec:TI}, we first introduce a Transmission Index that quantifies communication effectiveness for any pair of probabilistic inference and data selection processes. 
In Section~\ref{sec:TD}, we make connection between this index and the Average Teaching Dimension, thereby connecting our measure of effectiveness with previous measures of efficiency.
In Sections~\ref{sec:coop}, we introduce cooperative inference based on previous research in human social learning \citep{shafto2008teaching,Shafto2014}, present a Cooperative Index by extending the Transmission index to the cooperative setting, and identify the condition that must be satisfied to achieve optimal communication. 
In Section~\ref{sec:discussion}, we conclude with implications for human, machine, and human-machine learning.

\section{\label{sec:TI} THE TRANSMISSION INDEX}

In this section we define Transmission Index to quantify communication effectiveness. Communication occurs between two agents, which we call a teacher and a learner. Here the teacher represents the process of selecting data to convey a particular concept, and the learner represents the inference process of interpreting the received data.
In a probabilistic setting, the effectiveness of communication is related to the probability that the learner's interpretation matches the teacher's intended concept.



\begin{definition}\label{LT}
Let $\concept$ be a \textit{concept} in a 
\textit{concept space} $\conceptSpace$. A \textit{data set space}, $\dataSpace$, is a collection of subsets of a given 
set of data points.  $\data \in \dataSpace$ is called a \textit{data set}. Further, let $\TT$ be the teacher's probability of selecting a data set $\data$ for communicating a given concept $\concept$ and $\LL$ be the learner's posterior for $\concept$ given data set $\data$. We denote the size of $\conceptSpace$ and $\dataSpace$ by $\conceptSize$ and $\dataSize$, respectively.
\end{definition}


When $\conceptSpace$ and $\dataSpace$ are both discrete, in matrix notation, we can form the row-stochastic \textit{learner's inference matrix}, $\LLmat \in [0,1]^{\dataSize \times \conceptSize}$, having elements $\LL$, and the column-stochastic \textit{teacher's selection matrix}, $\TTmat \in [0,1]^{\dataSize \times \conceptSize}$, having elements $\TT$. 
As it is possible that there exist data sets (or concepts) whose probability of being selected is zero, here we allow a row (or column) stochastic matrix to have zero rows (or zero columns).

\begin{definition}\label{def:TI} 
The \textit{Transmission Index} ($\TI$) is defined as
\[\TI(\LLmat,\TTmat) = \frac{1}{\conceptSize} \sumhj \sumxi \LLmat_{i,j} \TTmat_{i,j}.\]
\end{definition}

Note that in the above definition, both $\conceptSize$ and $\dataSize$ can either be finite or countably infinite.
$\TI$ is well-defined when $\dataSize$ is countably infinite because  
$C_j:=\sumxi \LLmat_{i,j} \TTmat_{i,j}$ still converges in this case. ($C_j\leq \sumxi \TTmat_{i,j}=1$ and is thus bounded above, and each $\LLmat_{i,j} \TTmat_{i,j}$ is non-negative.)
When $\conceptSize$ is countably infinite, $\TI$ should be interpreted as a limit. See remark \ref{TI_infinite} for more detail.


In connection to channel coding in information theory, the learner's inference process is analogous to the decoding process, and the teacher's data selection process can be thought of as the combination of choosing the code words and passing them through a noisy channel, which makes the transmitted signals stochastic. Therefore, the Transmission Index can be related to channel capacity and the mutual information between the code words and the observations. These relationships deserve a full treatment that is outside the scope of this paper.


Now we give a few examples to show that $\TI$ captures how well on average a concept in a given concept space can be communicated with a given data set space.
Also, note that in the case where $\conceptSpace$ and $\dataSpace$ are clear from the context, we represent $\TI(\LLmat, \TTmat)$ simply by $\TI$.

\begin{example}
Let $\dataSize=\conceptSize=2$. Consider this teacher's selection matrix, 
$\TTmat =\begin{pmatrix}
    1       & 0  \\
    0       & 1  
\end{pmatrix}$, and these three learner's inference matrices, 
$\LLmatit{a} =\begin{pmatrix}
    1       & 0  \\
    0       & 1  
\end{pmatrix}$,
$\LLmatit{b} =\begin{pmatrix}
    0       & 1  \\
    1       & 0  
\end{pmatrix}$, and
$\LLmatit{c} =\begin{pmatrix}
    \sfrac{1}{2}       & \sfrac{1}{2}  \\
    \sfrac{1}{2}       & \sfrac{1}{2}  
\end{pmatrix}$.

In the first case (a), $\TI(\LLmatit{a},\TTmat) = 1$, because the concept that the teacher intends to teach through a certain data set matches perfectly what the learner would infer given that data set.  
In the second case (b), $\TI(\LLmatit{b},\TTmat) = 0$, because the concept that the teacher intends to teach through a certain data set leads the learner to infer the other concept with certainty. 
In the last case (c), $\TI(\LLmatit{c},\TTmat) = \frac{1}{2}$. Here the learner's inference is ambiguous, and $\TI$ captures that. In summary, $\TI$ captures the expected probability that the learner will interpret the teacher's intention correctly.
\end{example}

\begin{proposition}\label{TI_range}
Suppose that $\conceptSize$ is finite  and $\dataSize$ is finite or countably infinite \footnote{Similar conclusion also holds when $\conceptSize$ is countable infinite. See remark \ref{TI_infinite} for more detail.}, then the range of the Transmission Index is $0 \leq \TI \leq 1$, and $\TI=1$ if and only if two conditions hold:
(i) $\LLmat_{i,j}=1$ if $\TTmat_{i,j}>0$ for all $i,j$,
and (ii) there is no zero column in $\LLmat$ and $\TTmat$.
Also, $\TI=1$ implies that $\dataSize \geq \conceptSize$, with equality achieved when $\LLmat$ and $\TTmat$ are the same permutation matrix. 
\end{proposition}

\begin{proof}
$\TI\geq0$ because $\TTmat$ and $\LLmat$ are stochastic matrices, and $\TI=0$ if and only if for any $i, j$, either $\LLmat_{i,j}=0$ or  $\TTmat_{i,j}=0$.

We show $\TI \leq 1$:

\begin{equation}
\label{Inequality_1}
\TI(\LLmat,\TTmat) = \frac{1}{\conceptSize} \sumhj \sumxi \LLmat_{i,j} \TTmat_{i,j}
\underset{(a)}{\leq} \frac{1}{\conceptSize} \sumhj \left( \sumxi \TTmat_{i,j}\right)  
\underset{(b)}{\leq} \frac{1}{\conceptSize} \sumhj 1=1.
\end{equation} 

Inequality (a) in \eqref{Inequality_1} becomes an equality if and only if condition (i) is satisfied. This is because in order for $\LLmat_{i,j}\TTmat_{i,j}$ = $\TTmat_{i,j}$, we need 
$(\LLmat_{i,j}-1)\TTmat_{i,j}=0$, and this implies that $\LLmat_{i,j}=1$ or $\TTmat_{i,j}=0$, for any $i,j$.
Inequality (b) in \eqref{Inequality_1} follows from $\TTmat$ being a column-stochastic matrix, and it becomes an equality if and only if condition (ii) is satisfied. 

Given that $\LLmat$ is a row-stochastic matrix, if $\LLmat_{i,j}=1$, then there is no other non-zero elements in row $i$. This means that there are at most $\dataSize$ elements with value one in $\LLmat$; hence, by condition (i) the number of non-zero elements in $\TTmat$ is at most $\dataSize$. Also, condition (ii) requires that the number of non-zero elements in $\TTmat$ be at least $\conceptSize$. Therefore, $\dataSize \geq \conceptSize$, with equality achieved if and only if $\TTmat$ has only one positive element for each column. Together with condition (i), this implies that $\LLmat$ has at least one element with value one in each column. Because $\LLmat$ is row-stochastic, this implies $\LLmat$ is a permutation matrix. Condition (i) also implies that if $\LLmat_{i,j}<1$, then $\TTmat_{i,j}=0$. Together with condition (ii), $\TTmat$ is the same permutation matrix.
\end{proof}

\begin{remark}\label{invariant}
It is clear that when $\conceptSize$ is finite, $\TI$ is invariant under 
joint row and column permutations of $\LLmat$ and $\TTmat$.
When $\size{\conceptSpace} = \size{\dataSpace}$ and $\TI=1$, row and column exchangeability implies that $\LLmat$ and $\TTmat$ can always be arranged into an identity matrix of order $\size{\conceptSpace}$.
\end{remark}

\begin{remark}\label{TI_infinite}
When $\conceptSize$ is countably infinite  and $\dataSize$ is either finite or countably infinite, the Transmission Index is generalized to:

\[\TI(\LLmat,\TTmat) := \lim_{n\to \conceptSize} \frac{1}{n} \sum_{j=1}^{n} \sum_{i=1}^{\dataSize} \LLmat_{i,j} \TTmat_{i,j},
\]
This can be interpreted as the following. Let 
$\displaystyle S_n=\sum_{j=1}^{n} \sum_{i=1}^{\dataSize} \LLmat_{i,j} \TTmat_{i,j}$, then
$\displaystyle \TI(\LLmat,\TTmat)=\lim_{n\to \infty} \frac{S_n}{n}$.
Intuitively, columns of $\TTmat$ provide an enumeration of concepts in $\conceptSpace$ and $\frac{S_n}{n}$ measures how well on average the first $n$ concepts can be communicated. Further, as all terms are non-negative, if the limit of $\{\frac{S_n}{n}\}$ exists, it does not depend on this particular enumeration. Therefore, naturally $ \TI(\LLmat,\TTmat)$ is defined to be the limit of $\{\frac{S_n}{n}\}$.

Regrading the existence of $\TI$, there are two cases. The proof of Proposition~\ref{TI_range} implies that $0\leq S_n\leq n$. One case is that the growth rate of $S_n$ is strictly slower than any linear function, then $\TI=0$. Otherwise, $\TI$ exists if and only if the sequence $\{C_j \}$ converges as $j\to \infty$, where $C_j =\sum_{i=1}^{\dataSize} \LLmat_{i,j} \TTmat_{i,j}$. These results provide a guideline on constructing $\LLmat$ and $\TTmat$ to guarantee the existence of $\TI$ when $\conceptSize$ is countably infinite. See Supplementary Material for full detail.

\end{remark}

In the rest of this paper, we assume that both $\conceptSize$ and $\dataSize$ are finite. Adopting the limit notations, similar analysis can be made when $\conceptSize$ and $\dataSize$ are countably infinite.





\section{\label{sec:TD}CONNECTION TO AVERAGE TEACHING DIMENSION}
In this section we make the connection between the Transmission Index and the Average Teaching Dimension. 
The Average Teaching Dimension is a variant of Teaching Dimension, a classic measure for quantifying the efficiency of teaching.
The Teaching Dimension is well-studied; it has formal connections with the VC Dimension \citep{goldman1995complexity} and has been analyzed for certain models in continuous concept space \citep{Liu2016} and in cooperative settings \citep{Zilles2008, Doliwa2014}. 
However, Teaching Dimension and these analyses assume a deterministic learning model and focus on efficiency rather than effectiveness. To make connection to the analysis of Teaching Dimension, we first extend the Transmission Index, a measure of effectiveness, to the \textit{Expected Teaching Dimension}, a measure of efficiency. Then we show that the Expected Teaching Dimension, which is well-defined for probabilistic knowledge transmission, is the same as the Average Teaching Dimension when knowledge transmission becomes deterministic.



The analyses of Teaching Dimension are typically couched in the concept learning framework. In this framework, a concept, $\concept$, is a function that maps an instance, $x$, to a label, $y$. By observing examples, pairs of $(x,y)$, the learner can rule out concepts that are not consistent with the examples. With this notation, we can define the Average Teaching Dimension:

\begin{definition}[Average Teaching Dimension] \label{ATD}
A concept $\concept \in \conceptSpace$ is {\em consistent} with a data set $\data$ if and only if for every data point 
$(x,y)\in \data$, $\concept(x) = y$. 
$\data \in \dataSpace$ is a {\em teaching set} for concept $\concept \in \conceptSpace$ if $\concept$, but no other concept in $\conceptSpace$, is consistent with $\data$. Let $\dataSpace^*\!(\concept) \subset \dataSpace$ be the collection of teaching sets in $\dataSpace$ for concept $\concept$.
The classical version of \textit{Average Teaching Dimension} \citep{Doliwa2014} is defined as follows:
First, for any $\concept \in \conceptSpace$, let 

\[
TD(\concept) = 
\begin{cases}
\begin{array}{ll}
     \infty & \mbox{if $\dataSpace^*\!(\concept)$  is empty}\\
     \mathrm{min}_{\data \in \dataSpace^*\!(\concept)} \size{\data} & \mbox{otherwise}
\end{array},
\end{cases}
\]where $\size{\data}$ is the size of the data set $D$. Then, the Average Teaching Dimension ($ATD$) for the concept space $\conceptSpace$ is 
\[
ATD(\conceptSpace) = \frac{1}{|\conceptSpace|} \sum_{\concept \in \conceptSpace} TD(\concept).
\]
\end{definition}

Expected Teaching Dimension extends the Transmission Index to incorporate data set size as follows:

\begin{definition} 
The \textit{Expected Teaching Dimension} ($ETD$) is defined as
\[
ETD(\conceptSpace) = \frac{\sumh \sumx \size{\data}\, \LL \TT}{\sumh\sumx\LL\TT}.
\]
\end{definition}

\begin{definition}\label{consistency_matrix}
Let $\Likmat \in [0,1]^{\dataSize\times\conceptSize}$ be a matrix, where the element $\Likmat_{i,j}$ represents the probability that $\concept_i$ is consistent with $\data_j$. We define $C \in \{0,1\}^{\dataSize\times\conceptSize}$ to be a \textit{consistency matrix}, where $C_{i,j} = 1$ if $\concept_i$ is consistent with $\data_j$ and $C_{i,j} =0$ otherwise. $C$ can be sampled from $\Likmat$ by treating $C_{i,j}$ as the outcome of a Bernoulli trial with parameter $\Likmat_{i,j}$.
\end{definition}

Probabilistic consistency is an extension of deterministic consistency in the face of uncertainty. There are at least two cases where uncertainty can arise. The first case is when there are multiple possible learning scenarios but the learner is uncertain about which scenario is active. In this context, the probability of being consistent is the proportion of scenarios in which the concept is consistent with the data. The second case is when there is measurement noise. In this context, the learner has uncertainty about the true value of the data and therefore is also uncertain about whether the data is consistent with the concept.

\begin{proposition}
\label{ATD_TI}
Let $\size{\conceptSpace}=\size{\dataSpace}=N$, and $C$ be a consistency matrix of size $N\times N$.
Let $\LLmat$ and $\TTmat$ be the the row-normalized and column-normalized matrices of $C$, respectively. Then, $ATD(\conceptSpace)$ is finite if and only if $\TI(\LLmat,\TTmat)=1$.
\end{proposition}

\begin{proof}


$ATD(\conceptSpace)$ is finite if and only if $TD(\concept)$ is finite for all $\concept\in\conceptSpace$.
Finite $TD(\concept)$ means that there is at least one teaching set $\data \in \dataSpace$ for $\concept$.
Let $\alpha_i \subset \{1,2,\dots, N\}$ be the index set for the teaching sets of $\concept_i$.
Because every $\data$ can only belong to at most one $\dataSpace^*(\concept_i)$, so $\alpha_i \subset \{1,\cdots,N\} \backslash \cup_{j\neq i} \alpha_j$ for every $i\in \{1,2,\dots, N\}$.
Further, because $\size{\dataSpace} = \size{\conceptSpace}$, this construction of $\alpha_i$ implies that if $\size{\alpha_i}>1$ for some $i$, then there must exist at least one $j\neq i$ with the property that $\size{\alpha_j}=0$.
However, because $TD(\concept_i)$ is finite, $\alpha_i$ cannot be an empty set for any $i$. Hence, $\size{\alpha_i}=1$ for all $i$. In particular, this implies that $C$ is a permutation matrix. Thus, $ATD(\conceptSpace)$ is finite if and only if $C$ is a permutation matrix. $C$ being a permutation matrix implies that $C=\LLmat=\TTmat$, which by Proposition~\ref{TI_range} is equivalent to $\TI(\LLmat,\TTmat)=1$.
\end{proof}

\begin{example}
If $\LLmat=\TTmat$ and is a permutation matrix, $C=\TTmat$.
As we proved in Proposition~\ref{ATD_TI}, $ETD$ is the same as $ATD$. 
\end{example}


\begin{example}
\label{TD_CI_Ex}
We give an example when $ETD$ is finite but $ATD$ is infinite in the probabilistic setting.
Let $|\conceptSpace| = |\mathcal D| =2$,
$\Likmat =\begin{pmatrix}
    1       & \sfrac{1}{2}  \\
    0       & \sfrac{1}{2}  
\end{pmatrix}$.
There are four possible consistency matrices that can be sampled from $\Likmat$: $C^{(a)} =\begin{pmatrix}
    1       & 0  \\
    0       & 0  
\end{pmatrix}$, $C^{(b)} =\begin{pmatrix}
    1       & 0  \\
    0       & 1  
\end{pmatrix}$, $C^{(c)} =\begin{pmatrix}
    1       & 1  \\
    0       & 0  
\end{pmatrix}$, $C^{(d)} =\begin{pmatrix}
    1       & 1  \\
    0       & 1  
\end{pmatrix}$.
For $C^{(a)}$, $C^{(c)}$ and $C^{(d)}$, the corresponding $ATD(\conceptSpace)$ is $\infty$, and for $C^{(b)}$ it is $\frac{|D_1|+|D_2|}{2}$. 
Let $\LLmat^{(*)}$ and $\TTmat^{(*)}$ be the row-normalized and column-normalized matrices of $C^{(*)}$, respectively, for $* \in \{a,b,c,d\}$. 
Then, $\TI(\LLmat^{(a)},\TTmat^{(a)}) = \frac{1}{2}$, 
$\TI(\LLmat^{(b)},\TTmat^{(b)}) = 1$,
$\TI(\LLmat^{(c)},\TTmat^{(c)}) = \frac{1}{2}$, 
and $\TI(\LLmat^{(d)},\TTmat^{(d)}) = \frac{5}{8}$,
with $ETD(\conceptSpace)=|\data_1|$, $\frac{|D_1|+|D_2|}{2}$, $|D_1|$, $\frac{3|D_1|+2|D_2|}{5}$, respectively. 
Thus, $ETD$ can be seen as an generalization of $ATD$ from scenarios of perfect transmission ($\TI=1$) to those of imperfect transmission ($0\leq\TI\leq 1$) as well.


In addition to uncertain learning scenarios and measurement noise, another way probabilistic transmission can enter is that $\Likmat$ represents the degree of consistency between data and hypotheses. In this case, a deterministic learner would need to make a decision on what the underlying true consistency matrix is. 
Consider $\Likmat =\begin{pmatrix}
    1       & \sfrac{1}{2}  \\
    0       & \sfrac{1}{2}  
\end{pmatrix}$ again.
A simple decision rule is to round $\Likmat_{i,j}$ up to $1$ if it exceeds a threshold and down to $0$ otherwise. 
This decision rule would result in either $C^{(a)}$ or $C^{(d)}$, both of which correspond to $ATD=\infty$. 
\end{example}


\section{\label{sec:coop} OPTIMAL COOPERATIVE INFERENCE}

The Transmission Index introduced in Section~\ref{sec:TI} assumes that the learner and teacher, or more abstractly, the inference process and the data selection process, are independent.
However, communication for the transmission of knowledge is often \textit{cooperative} (e.g., in pedagogy \citep{Eaves2016a} and conversations \citep{Kao2014}). Here, cooperation implies that the teacher's selection of data depends on what the learner is likely to infer and vice versa.
In this section, we formalize \textit{cooperative inference}, which captures this inter-dependency between the two processes of inference and selection and has been proposed as a model of human language and teaching \citep{Kao2014,shafto2008teaching,Shafto2014}.
It can be seen as a way to map one common convention to another one that is more effective 
at transmitting knowledge without \textit{a priori} agreement on the encoding of data-concept pairs \citep{Zilles2008}.
We define \textit{Cooperative Index} as a measure of communication effectiveness in the cooperative setting by applying the Transmission Index to cooperative inference. 
Then, we provide proofs regarding the form of the shared likelihood matrix required to maximize the cooperative index and hence optimize cooperative inference. 

 
\begin{definition}[Cooperative inference]\label{CI}
Let $\data\in\dataSpace$ and $\concept\in\conceptSpace$. We define cooperative inference as a system of two equations:
\begin{subequations}
\begin{align}
\LL &= \frac{\TT \LLpri}{\LLmar}
\label{eq:L}\\
\TT &= \frac{\LL \TTpri}{\TTmar},
\label{eq:T}
\end{align}
\label{eq:LT}
\end{subequations}
where $\LL$ and $\TT$ are  defined in Definition~\ref{LT}; 
$\LLpri$ is the learner's prior of $\concept$; 
$\TTpri$ is the teacher's prior of selecting $\data$;
$\LLmar = \sumh \TT \LLpri$ is the normalizing constant for $\LL$; 
and $\TTmar = \sumx \LL \TTpri$ is normalizing constant for $\TT$. 
\end{definition}

The cooperative inference equations in \eqref{eq:LT} can be solved using fixed-point iteration \citep{shafto2008teaching,Shafto2014}: First, define an initial likelihood\footnote{This is the \textit{shared likelihood} and \textit{common convention} mentioned in the beginning of this section.}, $\TT=\Lik$, for the first evaluation of \eqref{eq:L}. Then, given $\LLpri$ and $\TTpri$, one can evaluate \eqref{eq:L}, use the resulting $\LL$ to evaluate \eqref{eq:T}, use the resulting $\TT$ to evaluate \eqref{eq:L}, and iterate this process until convergence. By symmetry, the iteration can also begin with \eqref{eq:T}.
This symmetry implies that the initial likelihood matrix, $\Likmat \in [0,1]^{\dataSize \times \conceptSize}$ with elements $\Lik$, can be an arbitrary non-negative matrix because it always gets appropriately normalized in the first iteration.

For the remainder of the paper, we assume that $P_{L_0}$ and $P_{T_0}$ are uniform distributions over $\conceptSpace$ and $\dataSpace$, respectively. In this case, the the fixed-point iteration of \eqref{eq:LT} depends only on $\Likmat$ and is simply the repetition of column and row normalization of $\Likmat$. Without loss of generality, we also assume that the iteration begins with \eqref{eq:L}. 

\begin{definition}
Let $\LLmatit{k}$ and $\TTmatit{k}$ be the matrices with elements $\LL$ and $\TT$, respectively, at the $k^\mathrm{th}$ iteration of \eqref{eq:LT}.
If the iteration of \eqref{eq:LT} converges, we define $\LLmatit{\infty}:=\lim_{k\rightarrow\infty}\LLmatit{k}$ and $\TTmatit{\infty}:=\lim_{k\rightarrow\infty}\TTmatit{k}$.
\end{definition}

\begin{definition}[Cooperative Index, $\CI$]
Given $\Likmat$ and assuming that the fixed-point iteration of \eqref{eq:LT} converges, we define the cooperative index as
\[ \CI(\Likmat) = \TI(\LLmatit{\infty},\TTmatit{\infty}) = \frac{1}{\conceptSize} \sumhj \sumxi \LLmatit{\infty}_{i,j} \TTmatit{\infty}_{i,j}.\]
\end{definition}

\begin{remark}
Similarly to $\TI$, $\CI$ is also well-defined as a limit when both $\conceptSize$ and $\dataSize$ are countably infinite, provided that the fixed-point iteration of \eqref{eq:LT} converges.
\end{remark}

We further assume that $\Likmat$ is a square matrix unless otherwise stated. Then, the iteration of \eqref{eq:LT} becomes the well-known Sinkhorn-Knopp algorithm, which provably converges under certain conditions by Sinkhorn's theorem \citep{Sinkhorn1967}. 
With this connection, we provide conditions under which optimal cooperative inference is achievable. 

\begin{definition}[Positive diagonal]
If $\Likmat$ is an $n \times n$ square matrix and $\sigma$ is a permutation of $\{1,\cdots,n\}$, then a sequence of \textit{positive} elements $\{\Likmat_{i,\sigma(i)}\}_{i=1}^n$ is called a \textit{positive diagonal}. If $\sigma$ is the identity permutation, the diagonal is called the \textit{main diagonal}. 
\end{definition}


\begin{theorem}[A simpler version of Sinkhorn's theorem \citep{Sinkhorn1967}]
\label{Sinkhorn}
Given any non-negative square matrix $\Likmat$ with at least one positive diagonal, $\LLmatit{k}$ and $\TTmatit{k}$ in the fixed-point iteration of \eqref{eq:LT} converges to the same doubly stochastic matrix, $\Likmatconverge$, which contains neither zero columns nor zero rows, as $k\rightarrow\infty$. 
\end{theorem}

\begin{proof}
Here we provide a sketch of the proof (see Supplementary Material for full detail). We pick one positive diagonal. First we show the product of all elements on that diagonal is positive and upper-bounded by 1 throughout the fixed-point iteration of \eqref{eq:LT}. Given uniform priors on both hypothesis and data set space, we then use the inequality of arithmetic and geometric means to prove that the product either stays the same or increase throughout the iteration. Finally, monotone convergence theorem of real numbers guarantees that the product will converge to its supremum, at which point $\LLmat$ and $\TTmat$ must have converged to the same doubly stochastic matrix.
\end{proof}

As is for $\TI$, if  $\Likmat$ is clear from the context, we denote $\CI(\Likmat)$ simply by $\CI$ for brevity. Now, we give two simple examples: The first demonstrates the fixed-point iteration of \eqref{eq:LT}; the second compares full cooperative inference with a special case known as machine teaching \citep{Zhu2015}.

\begin{example} \label{ex:coop_inf1}
Let
$\Likmat = \begin{pmatrix}
1 & 1\\
0 & 1
\end{pmatrix}$, then $\LLmatit{k}=\begin{pmatrix}
1-\frac{1}{2k}  & \frac{1}{2k}\\
0   & 1
\end{pmatrix}$ and $ \TTmatit{k}=\begin{pmatrix}
1  & \frac{1}{2k+1}\\
0   & 1 - \frac{1}{2k+1}
\end{pmatrix}$.
Notice that zero elements remain zero throughout the iteration process, but non-zero elements may converge to zero.
Since $\LLmatit{\infty}$ and $\TTmatit{\infty}$ are both the identity matrix, $\CI=1$. In contrast, after one iteration of \eqref{eq:LT}, $\LLmatit{1} = \begin{pmatrix}
\sfrac{1}{2} & \sfrac{1}{2}\\
0 & 1
\end{pmatrix}$, $\TTmatit{1} = \begin{pmatrix}
1 & \sfrac{1}{3}\\
0 & \sfrac{2}{3}
\end{pmatrix}$, and $\TI(\LLmatit{1},\TTmatit{1})$ is only $\frac{2}{3}$. Similarly, for any $k$, $\TI(\LLmatit{k},\TTmatit{k})<1$. Thus, cooperative inference increases the effectiveness of communication.
\end{example}

\begin{example}\label{ex:mt} In this example we apply $\TI$ and $\CI$ to machine teaching in a simple setting. Following \cite{Liu2016}, consider a version-space learner who is trying to learn a threshold classifier $h_\theta$, $\theta \in \{1,2,3\}$. For $x \in \{0,1,2,3\}$, $h_\theta$ returns $y=-$ if $x < \theta$ and $y=+$ if $x \geq \theta$. Assume a teacher provides training set $D=\{(x_1, y_1),(x_2,y_2)\}$ and the learner assigns the same likelihood to all concepts that are consistent with the data; then, the learner's inference matrix is: 
\[\LLmat = \begin{tabular}{ |c|c|c|c|c| } 
        \hline
        $\{x_1,y_1,x_2,y_2\}\backslash h_\theta$ & $h_1$ & $h_2$ & $h_3$ \\ \hline
        $\{0,-,1,+\}$ & 1 & 0 & 0 \\ \hline
        $\{0,-,2,+\}$ & 1/2 & 1/2 & 0 \\ \hline
        $\{0,-,3,+\}$ & 1/3 & 1/3 & 1/3 \\ \hline
        $\{1,-,2,+\}$ & 0 & 1 & 0 \\ \hline
        $\{1,-,3,+\}$ & 0 & 1/2 & 1/2 \\ \hline
        $\{2,-,3,+\}$ & 0 & 0 & 1 \\ \hline        
    \end{tabular}\;.\]
Following \cite{Liu2016}, machine teaching chooses data that maximize the likelihood for the learner to infer the correct hypothesis. Note that this way of teaching can be considered as a special case of cooperative inference: The teacher selects data by maximizing $\PP{L}{\concept|\data}$ rather than sampling in proportion to the probability, and the learner does not reason about the teacher's selection and thus only the first step of the recursive cooperative inference is executed (see \ref{eq:L} and \ref{eq:T}).
Machine teaching will choose data sets $\{0,-,1,+\}$, $\{1,-,2,+\}$, and $\{2,-,3,+\}$ for $h_1$, $h_2$, and $h_3$, respectively, with probability $1$. Let machine teaching's data selection matrix be $\TTmat^{(mt)}$. The effectiveness of machine teaching can then be quantified by $\TI(\LLmat,\TTmat^{(mt)})$, which is $1$ in this case. Depending on the concept space and data set space, machine teaching's effectiveness is not always perfect. For example, if the learner's inference matrix consists of only the first three rows of $\LLmat$, $\TI$ for machine teaching becomes $0.611$.

\end{example}

Given the cooperative index, which quantifies the effectiveness of transmission for cooperative inference, we can prove conditions under which $\Likmat$ maximizes $\CI$:


\begin{definition}
A square matrix is \textit{triangular} if it has a positive main diagonal, and has only zeros below (upper-triangular) or above (lower-triangular) the main diagonal. 
\end{definition}


\begin{theorem}[Representation theorem for cooperative inference]\label{optCI}
Let $\Likmat$ be a nonnegative square matrix with at least one positive diagonal, then
the following statements are equivalent:
\begin{enumerate}[(a)]
    \item The cooperative index is optimal, i.e., $\CI(\Likmat)=1$;
    \item $\Likmat$ has exactly one positive diagonal;
    \item $\Likmat$ is a permutation of an upper-triangular matrix.    
\end{enumerate}
\end{theorem}

\begin{proof}
From Proposition \ref{TI_range} we know that $\CI(\Likmat)=\TI(\Likmatconverge,\Likmatconverge) = 1$ if and only if $\Likmatconverge$ is a permutation matrix.
Since elements of $\Likmat$ that lie in a positive diagonal do not tend to zero during cooperative inference \citep{Sinkhorn1967} (i.e., if $\Likmat_{i,j}\neq 0$ lies in a positive diagonal, then $\Likmatconverge_{i,j}\neq 0$), $\Likmatconverge$ is a permutation matrix if and only if $\Likmat$ has exactly one positive diagonal. So we have $(a)\Longleftrightarrow (b)$. $(b)\Longleftrightarrow (c)$ is a fact of linear algebra which can be proved by induction on the dimension $n$ of $\Likmat$ (see Supplementary Material for full detail).
\end{proof}


\begin{remark}
Let $C$ be a consistency matrix of size $N\times N$ as in Definition~\ref{consistency_matrix}. Suppose that $C$ is a permeation of an upper-triangular matrix, then $\CI(C)=1$. Together with Proposition~\ref{ATD_TI}, we have that the Average Teaching Dimension of the corresponding concept space $\conceptSpace$ is finite at the convergence of the cooperative inference iteration, but is infinite before that (unless $C$ is a permutation matrix). 

\end{remark}

Theorem~\ref{optCI} shows that in order to achieve optimal cooperative inference and thereby effective knowledge accumulation, the shared inductive bias should be one that constraints the form of $\Likmat$ to be upper triangular (or a permutation thereof). 
This in turn constraints the learner's likelihood function such that it applies zero probability to particular data-concept relationships.
Below, we show an example of using $\CI$ to investigate the form of the likelihood that leads to optimal transmission effectiveness. 

\begin{figure*}
\vspace{.3in}
	\centering
	\includegraphics[width=1\textwidth]{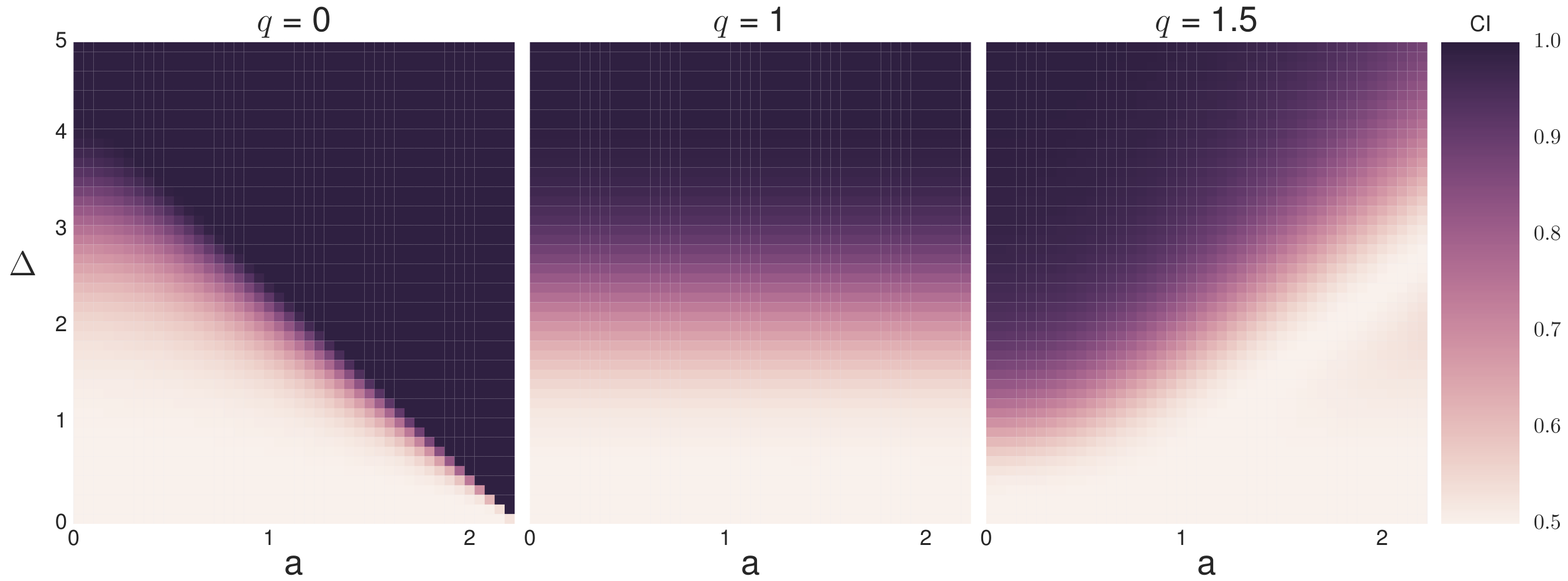}
\vspace{.3in}
	\caption{Comparison of $\CI$ across three different error likelihood functions (based on $q$-Gaussian distributions with different values of $q$) in polynomial regression. Each of the plots illustrates how $\CI$ varies as a function of the parameters $a$ and $\Delta$ that specify different data set spaces (see main text and Supplementary Material for detail). We find that only having a compact error distribution, i.e., $q=0$, results in optimal $\CI$ for all settings of $\Delta$, which corresponds to the signal strength in the data.}
	\label{fig:ci_example_plots}
\end{figure*} 
  
\begin{example}\label{poly}
Consider polynomial regression. 
In order to have a triangular $\Likmat$, the likelihood must have finite support.
We explore the behavior of $\CI$ under different likelihood functions, ranging from fat-tailed to compact.
In particular, we explore the conditions under which the different distributions lead to optimal $\CI$.

Let $\{x_i\}_{i=1}^6=\{-1,-1,0,0,1,1\}$ and $\{y_i\}_{i=1}^6=\{a, -a, \Delta+a, \Delta-a, a, -a\}$.
The quantity $\Delta/a$ can be viewed as the signal-to-noise ratio for a second-order polynomial.
Let $\dataSpace=\{\data_1, \data_2\}$, where $\data_1=\{x_1,\ldots,x_4,y_1,\ldots,y_4\}$ and $\data_2=\{x_1,\ldots,x_6,y_1,\ldots,y_6\}$.
Let $\conceptSpace=\{\concept_1,\concept_2\}$, where $\concept_i$ is a polynomial of order $i$ with a likelihood function that defines the assumed noise distribution.
The likelihood function is a q-Gaussian $N_q(z;\mu)$ with unit variance \citep{tsallis2009introduction}.
We construct the $\Likmat$ via maximum likelihood as a function of $\Delta$ and $a$ for $q=\{0,1,1.5\}$. For each value of $q$, we first find the maximum-likelihood estimate of $\concept_i$ to $\data_j$, then assign $\Likmat_{i,j}$ the likelihood produced by that estimate (see Supplementary Material for more details). Having obtained these $\Likmat$ matrices, we iterate them according to \eqref{eq:LT} to explore the behavior of $\CI$.

In Figure~\ref{fig:ci_example_plots} we show the phase diagrams of $\CI$ for the three q-Gaussian distributions, which correspond to a compact ($q=0$), normal ($q=1$), and fat-tailed ($q=1.5$) distribution. This result shows that when the error likelihood is a compact distribution, there exists at least one setting of $a$ such that $\CI=1$ for all $\Delta>0$. This is not the case when the error likelihood has infinite support, i.e., $q=1$ or $q=1.5$. As suggested by Theorem~\ref{optCI}, modeling choices that yield $\Likmat$'s that are closer to triangular, such as compact likelihood functions, can produce optimal cooperative inference. This illustrates a simple modeling choice that
allows a small set of examples to uniquely identify different parameterizations of the model. It is in this sense that optimization of the Cooperative Index may foster explainability and interpretability---by allowing small sets of examples to uniquely map to underlying parameterizations of the model, without requiring that the maps between hypotheses and data be bijective. 

\end{example}

\section{\label{sec:related}RELATED WORK}

As briefly discussed in Example~\ref{ex:mt}, machine teaching is a close cousin of cooperative inference in that both aim to choose good data to convey a target concept. Machine teaching can be thought of as performing only one step of the cooperative inference iteration and choosing deterministically the best choice available. In this setting, \cite{Liu2016} has derived the Teaching Dimension for linear learning models and discussed the connections to VC Dimension. For simpler version-space learner models, \cite{Doliwa2014} has made formal connections between the Teaching Dimension, VC Dimension, and sample compression in the iterative setting, and  \cite{searcy2016cooperative} investigated the representational implications of deterministic cooperation. 
These differ from CI in that they assume deterministic, rather than probabilistic, inference.

Furthermore, since cooperative inference is implemented via
the Sinkhorn-Knopp algorithm, many connections stem from the body of work relating to Sinknorn's scaling (see \cite{idel2016review} for review). 
To give a few examples, on the theoretical front, Sinkhorn's theorem has been analyzed with geometric interpretation \citep{dykstra1985iterative}, in a convex programming formulation \citep{macgill1977theoretical,krupp1979properties}, and as an entropy minimization problem with linear constraint \citep{brown1993order}.
On the application side, Sinkhorn's theorem has been applied to modelling transportation \citep{de1994modelling}, designing condition numbers \citep{benzi2002preconditioning}, and ranking webpages \citep{knight2008sinkhorn}.


\section{\label{sec:discussion}DISCUSSION}



Cooperative inference is central to human and machine learning. Previous work has introduced numerous accounts of the role of cooperation in learning and applied these across a host of problems in human and machine learning; however, to date, there has been no account of when or why we should expect cooperative inference to outperform simple learning. Building on prior models of cooperation from cognitive science of language and learning and demonstrating connections to models of machine teaching, we investigate this question. We introduced the Transmission and Cooperative Indices, which are metrics for the effectiveness of inference in standard learning and cooperative learning settings, respectively. We connect the Transmission Index with prior measures of efficiency in deterministic settings, namely, Teaching Dimension, and prove a representation theorem stating the conditions under which cooperation can yield optimally effective inference. We demonstrate how this model informs modification of a standard model of learning to ensure optimal cooperative transmission of the model class via a small subset of the data. 

Beyond human learning, where this work provides foundational theory to inform accounts of human cognitive development, language, and cultural evolution, this work has strong implications for development of machine learning models that are designed for explainability and interpretability. Implicit in these is the existence of a shared goal, and cooperation is the natural formalization of this. Whereas models necessarily encode inferences about data in an internal language, and those internal languages may take many different forms depending on the task or domain, data provide a general purpose language in which inferences can be encoded to and decoded from. The promise of this work is that it provides an overarching framework for thinking about how to engineer models that are not only predictively accurate, but also understood well enough to be deployed correctly. 

There are a number of practical and theoretical reasons to be concerned with the explanability of machine learning and AI algorithms. Practical reasons are related to algorithms' use in industry, for example, to decide who will get loans or determine prison sentences. Human intelligibility to ensure the algorithms are not simply propagating race, gender or other biases as well as to satisfy recent legal standards is necessary (see recent EU laws related to a right to an explanation; \cite{goodman2016european}). Theoretical reasons are highlighted by the adversarial images that illustrate how little we understand the workings of deep learning (and probably other classes of) models. Our paper presents theoretical results upon which we may develop systems that are designed to be explainable by building models that adopt the structural constraints necessary to ensure optimal cooperative inference.

\section{Acknowledgements}

This material is based on research sponsored by the Air Force Research Laboratory and DARPA under agreement number FA8750-17-2-0146 to P.S. and S.Y. The U.S. Government is authorized to reproduce and distribute reprints for Governmental purposes notwithstanding any copyright notation thereon. This research was also supported by NSF SMA-1640816 to PS.

\bibliographystyle{plainnat} 


\newpage


\section{Supplementary Material: Optimal Cooperative Inference}




This supplementary material presents the additional details and proofs associated with the main paper.

\subsection{Details of Remark~\ref{TI_infinite}}
Suppose that $\conceptSize$ is countably infinite. Let $\mathbf{A}=(\LLmat_{i,j} \TTmat_{i,j})_{\dataSize \times \conceptSize}$ be the matrix obtained from $\LLmat$ and $\TTmat$ by element-wise multiplication. Denote the sum of elements in the $j$-th column of $\mathbf{A}$ by $C_j$. Then $S_n=\sum_{j=1}^{n} C_j$ is the sum of elements in the first $n$ columns of $\mathbf{A}$. Note that $\displaystyle 0\leq C_j=\sum_{i=1}^{\dataSize} \LLmat_{i,j} \TTmat_{i,j} \leq \sum_{i=1}^{\dataSize} \TTmat_{i,j} =1 $ and so $0\leq S_n \leq n$. Therefore, for any $j, n$, both $C_j$ and $S_n$ exist, and $\{\frac{S_n}{n}\}_{n=1}^{\infty}$ is a well-defined sequence whose limit is then called $\TI$.  

Regrading the existence of $\TI$, there are two cases. 

Case 1: The growth rate of $S_n$ is strictly slower than any linear function. Thus, for any $k>0$, there exists an integer $N(k)>0$ (depends on $k$) such that $S_n< k \cdot n$ for any $n>N(k)$. 
Then for any $k>0$, the following holds:
$$0 \leq \TI=\lim_{n\to \infty} \frac{S_n}{n}\leq \lim_{n\to \infty}\frac{k\cdot n}{n}=k.$$
Thus, $\TI=0$.

Case 2: If the growth rate of $S_n$ is not strictly slower than linear functions, then $\TI$ exists if and only if the sequence $\{C_j \}$ converges as $j\to \infty$. Suppose that $\{C_j \}$ converges to $k$. Then for any $\epsilon>0$, there exists an integer $N(\epsilon)$ such that $|C_m-k|< \epsilon$ for any $m>N(\epsilon)$. Therefore, for $n$ sufficiently large, 

$$|\frac{S_n}{n}-k|=|\frac{S_n-n\cdot k}{n}|=|\frac{S_{N}-N\cdot k}{n}+\frac{\sum_{j=N}^n C_j-k}{n-N}|\leq |\frac{S_{N}-N\cdot k}{n}|+\epsilon\leq \epsilon'.$$
Thus, $\TI$ exists. Similarly the other direction also holds.

Moreover, when $\TI$ exists, Proposition~\ref{TI_range} can also be generalized. $0 \leq S_n\leq n$ implies that the range of $\TI$ is $[0,1]$, and $\TI=1$ if and only if $C_j$ converges to $1$.

\subsection{Proof of Theorem \ref{Sinkhorn}}

For convenience, we first write the fixed-point iteration of \eqref{eq:LT} explicitly in vector form.
We denote the matrix with elements $\LL$ by $\LLmat \in [0,1]^{\dataSize \times \conceptSize}$, the matrix with elements $\TT$ by $\TTmat \in [0,1]^{\dataSize \times \conceptSize}$, and the matrix with elements $\Lik$ by $\Likmat \in [0,1]^{\dataSize \times \conceptSize}$. 
Further, denote the vectors consisting of $\LLpri$ and $\TTmar$ by $\LLvecpri, \TTvecmar \in [0,1]^{\conceptSize \times 1}$, vectors consisting of $\TTpri$ and $\LLmar$ by $\TTvecpri, \LLvecmar \in [0,1]^{\dataSize \times 1}$, respectively.
Given $\LLvecpri$, $\TTvecpri$, and $\Likmat$, the fixed-point iteration of the cooperative inference equations can be expressed as:
\begin{subequations}
\begin{alignat}{3}
\LLit{1} &= \frac{\Lik\LLpri}{\LLitmar{1}}
&&\;\;\Longleftrightarrow\;\;
\LLmatit{1} = \Diag{\frac{1}{\Likmat\LLvecpri}} \Likmat \LLmatpri
\label{eq:learn_1}\\
\TTit{k+1} &= \frac{\LLit{k+1}\TTpri}{\TTitmar{k+1}}
&&\;\;\Longleftrightarrow\;\;
\TTmatit{k+1} = \TTmatpri \LLmatit{k+1} \Diag{\frac{1}{\TTvecmarit{k+1}}}
\label{eq:teach_k}\\
\TTitmar{k+1} &= \sumx \LLit{k}\TTpri     
&&\;\;\Longleftrightarrow\;\;
\TTvecmarit{k+1} = \transpose{(\LLmatit{k+1})}\, \TTvecpri 
\label{eq:teach_nor}\\
\LLit{k+1} &= \frac{\TTit{k}\LLpri}{\LLitmar{k+1}}
&&\;\;\Longleftrightarrow\;\;
\LLmatit{k+1}= \Diag{\frac{1}{\LLvecmarit{k+1}}} \TTmatit{k} \LLmatpri
\label{eq:learn_k}\\
\LLitmar{k+1} &= \sumh \TTit{k}\LLpri
&&\;\;\Longleftrightarrow\;\;
\LLvecmarit{k+1} = \TTmatit{k} \LLvecpri,
\label{eq:learn_nor}
\end{alignat}
\label{eq:CI}
\end{subequations}
where $k$ denotes the iteration step; $\Diag{\mathbf{z}}$ denotes the diagonal matrix with elements of the vector $\mathbf{z}$ on its diagonal; and $\frac{1}{\mathbf{z}}$ denotes element-wise inverse of vector $\mathbf{z}$.

Note that \eqref{eq:teach_k} and \eqref{eq:teach_nor} are the operations to column normalize $\TTmatpri\LLmatit{k}$, and \eqref{eq:learn_k} and \eqref{eq:learn_nor} are the operations to row normalize $\TTmatit{k}\LLmatpri$. Zero rows in $\LLmatit{k}$ and zero columns in $\TTmatit{k}$ are fixed throughout the iteration of \eqref{eq:CI} if they exist. 
This is equivalent to removing the zero rows and zero columns of $\Likmat$ for \eqref{eq:CI} and inserting them back at convergence or when the iteration is stopped. 

Now we provide a version of the proof using the notations introduced in the paper. The original proof can be found in \citep{Sinkhorn1967}. Remember that $\LLvecpri$ and $\TTvecpri$ are assumed to be uniform.

\begin{proof}
Let $\permu$ be a permutation of $\{1,\cdots,n\}$ that makes $\{\Likmat_{i,\permu(i)}\}_{i=1}^n$ a positive diagonal. Define
\begin{equation*}
  e^{(k)}:=\prod_{i=1}^n \LLmatit{k}_{i,\permu(i)};\quad f^{(k)}:=\prod_{i=1}^n \TTmatit{k}_{i,\permu(i)}.
\end{equation*}
Applying \eqref{eq:learn_1},
$\LLmatit{1}$ is a row-stochastic matrix, and $\{\LLmatit{1}_{i,\permu(i)}\}_{i=1}^n$ is a positive diagonal, hence $e^{(1)}$ is positive.
Also, by applying (\ref{eq:teach_k}),
\begin{equation}
  f^{(1)} =\prod_{i=1}^n\TTmatit{1}_{i,\permu(i)} =\prod_{i=1}^n\left(\TTvecpri_i\frac{\LLmatit{1}_{i,\permu(i)}}{\TTvecmarit{1}_{\permu(i)}}\right)
  =\frac{e^{(1)}}
  {n^n \prod_{i=1}^n \TTvecmarit{1}_{\permu(i)}}
  =\frac{e^{(1)}}
  {n^n\prod_{i=1}^n\TTvecmarit{1}_i}.
  \label{eq:fe}
\end{equation}
By the inequality of arithmetic and geometric means,
$\left(\prod_{i=1}^n\TTvecmarit{1}_i\right)^\frac{1}{n}\leq\frac{1}{n}\sum_{i=1}^n\TTvecmarit{1}_i$. Also,
$\LLmatit{1}$ is a row-stochastic matrix and we assumed uniform prior on data set space, and hence, by (\ref{eq:teach_nor})
\begin{equation}
  n^n\prod_{j=1}^n\TTvecmarit{1}_j \leq
  \left(\sum_{j=1}^n\TTvecmarit{1}_j\right)^n
  =\left(\sum_{i=1}^n\sum_{j=1}^n\TTvecpri_j\LLmatit{1}_{i,j}\right)^n
  =\left(\frac{1}{n}\sum_{i=1}^n\sum_{j=1}^n\LLmatit{1}_{i,j}\right)^n
  =1.
\label{eq:amgm}
\end{equation}
The equality in \eqref{eq:amgm} is achieved if and only if $\TTvecmar=\left(\frac{1}{n},\dots,\frac{1}{n}\right)$, or equivalently, $\LLmatit{1}$ being a doubly stochastic matrix. Because $f^{(1)}$ is the product of $n$ values between 0 and 1,
\begin{equation}
0 < e^{(1)} \underset{(a)}{\leq} f^{(1)} \underset{(b)}{\leq} 1,
\end{equation}
with equality in (a) if and only if $\LLmatit{1}$ is a doubly stochastic matrix, and equality in (b) if and only if $\LLmatit{1}$ is a permutation matrix. 
Applying the same logic to equations (\ref{eq:learn_k}) and (\ref{eq:learn_nor}), we have
\[0 < f^{(1)} \underset{(c)}{\leq} e^{(2)} \underset{(d)}{\leq} 1,\]
with equality in (c) if and only if $\TTmatit{1}$ is a doubly stochastic matrix, and equality in (d) if and only if $\TTmatit{1}$ is a permutation matrix. Repeating this argument, we get  the increasing sequence 
\[0 < e^{(1)} \leq f^{(1)} \leq e^{(2)} \leq f^{(2)} \leq \dots \leq 1.\]
Monotone convergence theorem of real numbers guarantees that this sequence converges to its supremum
\[\lim_{k\to\infty} e^{(k)} = \lim_{k\to\infty} f^{(k)} = \sup\{\mathbf{e},\mathbf{f}\}.\]
Asymptotically, $e^{(k)}=f^{(k)}=e^{(k+1)}$; therefore, $\LLmatit{k}$ and $\TTmatit{k}$ are both doubly stochastic matrices. Because doubly stochastic matrices are stable under row and column normalization,  $\LLmat$ and $\TTmat$ converge to the same doubly stochastic matrix,
\[ \Likmatconverge := \lim_{k\to\infty} \LLmatit{k} = \lim_{k\to\infty} \TTmatit{k}.\]
\end{proof}

\subsection{Proof of Theorem \ref{optCI}}
\begin{proof}

(1) $(a)\Longleftrightarrow (b) $: We first prove that (a) $\CI(\Likmat)=1$, and  (b) \textit{$\Likmat$ has exactly one positive diagonal,} are equivalent. Since $\Likmat$ is an $n\times n$ nonnegative matrix with at least one positive diagonal, Theorem \ref{Sinkhorn} guarantees that the iteration of equation set (\ref{eq:CI}) converges to a doubly stochastic matrix, $\Likmatconverge$.
According to Birkhoff–von Neumann theorem \citep{Birkhoff1946, Von1953}, 
there exist $\displaystyle \theta _{1},\ldots ,\theta _{k}\in (0,1]$ with $\sum_i \theta_i=1$ and distinct permutation matrices $\displaystyle P_{1},\ldots ,P_{k}$ such that $\Likmatconverge =\theta _{1}P_{1}+\cdots +\theta _{k}P_{k}$. 
 To simplify, we adopt the \textit{inner product} notation between matrices: $A\boldsymbol{\cdot} B= \sum_{i,j} A_{i,j}B_{i,j}$, for any two $n\times n$ square matrices $A$ and $B$. 
 Then the following holds:

 $$\CI=\TI(\Likmatconverge,\Likmatconverge) \underset{(I)}{=} \frac{1}{n} \Likmatconverge \boldsymbol{\cdot} \Likmatconverge \underset{(II)}{=}\frac{1}{n} (\sum_{i} \theta_{i} P_i) \boldsymbol{\cdot} (\sum_{j} \theta_{j} P_j) \underset{(III)}{=}\frac{1}{n} \sum_{i,j} \theta_{i}\theta_{j}P_i \boldsymbol{\cdot} P_j.$$
 
Equality (I) comes from rewriting $\TI$ in the inner product notation. Equality (II) comes from substituting $\Likmatconverge$ by its Birkhoff–von Neumann decomposition. Equality (III) comes from distribution.
 
Further, as permutation matrices,  $P_i\boldsymbol{\cdot} P_j \leq n$, and the equality holds if and only if $P_i=P_j$. So we have 
$$
\CI(\Likmat) =\frac{1}{n} \sum_{i,j} \theta_{i}\theta_{j}P_i \boldsymbol{\cdot} P_j\underset{(IV)}{ \leq} \frac{1}{n} \sum_{i,j} \theta_{i}\theta_{j} n = \sum_{i,j} \theta_{i}\theta_{j} = (\sum_i \theta_{i}) \times (\sum_j \theta_{j})=1.
$$

\noindent The equality in (IV) holds if and only if $P_i=P_j$ for any $i,j$. Note that $\displaystyle P_{1},\ldots ,P_{k}$ are distinct, i.e., $P_i\neq P_j$ when $i\neq j$. So the equality in (IV) is achieved precisely when $k=1$ and $\Likmatconverge=P_1$. Hence, $\CI(\Likmat)$ is maximized if and only if $\Likmatconverge$ is a permutation matrix. 


We then prove that $\Likmatconverge$ is a permutation matrix if and only if $\Likmat$ has exactly one positive diagonal. 
This follows from this claim, \textbf{Claim $(1)$}: elements of $\Likmat$ that lie in a positive diagonal do not tend to zero during the cooperative inference iteration \citep{Sinkhorn1967} (i.e., if $\Likmat_{i,j}\neq 0$ lies in a positive diagonal, then $\Likmatconverge_{i,j}\neq 0$). Claim $(1)$ implies that $\Likmatconverge$ and $\Likmat$ have the same number of positive diagonals. 
Further, note that a doubly stochastic matrix has exactly one diagonal if and only it is a permutation matrix. So as a doubly stochastic matrix, $\Likmatconverge$ is a permutation matrix if and only if $\Likmat$ has exactly one positive diagonal. Thus, $\CI$ is maximized if and only if $\Likmat$ has exactly one positive diagonal.  

To complete the proof for $(a) \Longleftrightarrow (b)$, we only need to justify Claim $(1)$. Note that the product of any positive diagonal converges to a positive number $\sup\{\mathbf{e},\mathbf{f}\}$ (shown in the proof for Theorem \ref{Sinkhorn}) and all elements on the positive diagonal is upper-bounded by 1 and lower-bounded by $\sup\{\mathbf{e},\mathbf{f}\}$. , elements on a diagonal of $\Likmat$ cannot converge to 0.



\vspace{1pc}

(2) $(b)\Longleftrightarrow (c)$: This follows immediately from a slightly more general claim below, where positive diagonals are generalized to non-zero diagonals (can have negative values).


\textbf{Claim (2)}: Let $A$ be an $n\times n$-square matrix (elements can be any real number). Then $A$ has exactly one non-zero diagonal (i.e., a diagonal with no zero element) if and only if $A$ is a permutation of an upper-triangular matrix. 

We now prove Claim (2). The if direction is clear since an upper-triangular matrix always has exactly one non-zero diagonal, which is its main diagonal. The only if direction is proved by induction on the dimension $n$ of $A$.
\vspace{1pc}

\textbf{Step 1---Induction basis}: When $n=2$, it is easy to check that any $2\times 2$ matrix with exactly one diagonal is either of the form 
$\begin{pmatrix}
a & b\\
0 & c
\end{pmatrix}$ or  $\begin{pmatrix}
a & 0\\
b & c
\end{pmatrix}$, where $a,c\neq 0$. So it is a permutation of an upper-triangular matrix.

\vspace{1pc}

\textbf{Step 2---Inductive step}: Suppose that the claim---an $n\times n$-square matrix $A$ has exactly one non-zero diagonal if and only if it is a permutation of an upper-triangular matrix---holds for any $n<N$. We need to show that the claim also holds when $n=N$.

The following notation will be used. Let $A$ be an $n\times n$-square matrix. $A_{i,j}$ denotes the element of $A$ at row~$i$ and column~$j$. $\widetilde{A} _{i,j}$ denotes the $(n-1)\times (n-1)$ sub-matrix obtained from $A$ by crossing out row~$i$ and column~$j$.

First, we will prove three handy observations.  

\textbf{Observation 1}: If $A$ has exactly one non-zero diagonal and $A_{i,j}\neq 0$, then $\widetilde{A} _{i,j}$ has at most one non-zero diagonal. In particular, if $A_{i,j}$ is on that non-zero diagonal, then $\widetilde{A} _{i,j}$ has exactly one non-zero diagonal.

\textit{Proof of Observation 1}: Suppose that $\widetilde{A} _{i,j}$ has more than one diagonal. Then these diagonals for $\widetilde{A} _{i,j}$ along with $A_{i,j}$ form different diagonals for $A$, which is a contradiction.

\textbf{Observation 2}: If $A$ has exactly one non-zero diagonal and $A$ has a row or a column with exactly one non-zero element, then $A$ is a permutation of an upper-triangular matrix.

\textit{Proof of Observation 2}: Suppose that $A$ has a column with exactly one non-zero element. Then by permutation, we may assume that it is the first column of $A$ and the only non-zero element in column $1$ is $A_{1,1}$. $A_{1,1}$ must be on the non-zero diagonal of $A$. Hence, according to observation 1, 
$\widetilde{A} _{1,1}$ is a $(N-1\times N-1)$-square matrix with exactly one non-zero diagonal. Then by the inductive assumption, we may permute $\widetilde{A}_{1,1}$ into an upper-triangular matrix. Note that each permutation of $\widetilde{A} _{1,1}$ induces a permutation of $A$. So there exist permutations that convert $A$ into $A'$ such that $A'_{i,j}=0$ when $j>1$ and $i>j$. Moreover, permutations that convert $A$ to $A'$ never switch column $1$ (row $1$) of $A$ with any other columns (rows). So $A'_{i, 1}=0$ for $i\neq1$, as $A_{1,1}$ is the only non-zero element in the first column of $A$. Thus, we have $A'_{i,j}=0$ when $i> j$, which implies that $A'$ is an upper-triangular matrix.

If $A$ has a row with exactly one non-zero element, then up to permutation, we may assume it is the last row of $A$ and the only non-zero element is $A_{N,N}$. Following similar argument as above, we may show that $\widetilde{A} _{N,N}$ can be arranged into an upper-triangular matrix by permutations. The corresponding permutations of $A$ will also convert $A$ into an upper triangular matrix. So observation 2 holds.

\textbf{Observation 3}: If the main diagonal of $A$ is the only non-zero diagonal of $A$, then
$A_{t_1,t_2}A_{t_2,t_3}\cdots A_{t_{k-1}, t_k}A_{t_k,t_1}= 0$ for any distinct $t_1,t_2,\dots, t_k$.


\textit{Proof of Observation 3}: Suppose that $A_{t_1,t_2}A_{t_2,t_3}\cdots A_{t_{k-1}, t_k}A_{t_k,t_1}\neq 0$. Then a different non-zero diagonal for $A$ other than the main diagonal is form by $\{A_{i,i}|i\neq t_1, \dots, t_k\}$ and $A_{t_1,t_2}, A_{t_2,t_3}, \cdots,  A_{t_{k-1}, t_k}, A_{t_k,t_1}$. 
\vspace{1pc}

Now back to the inductive step. Suppose that $A$ is an $N\times N$-square matrix with exactly one non-zero diagonal.  By permutation, we may assume that the main diagonal of $A$ is the only non-zero diagonal. In particular, $A_{1,1}\neq 0$. 
According to Observation~1, $\widetilde{A} _{1,1}$ has exactly one non-zero diagonal and so can be arranged into an upper-triangular matrix by permutations. The corresponding permutations convert $A$ into a new form, denoted by $A^1$, with the property that $A^1_{i,j}=0$ when $j>1$ and $i>j$. In particular, $A^1_{Nj}=0$ when $j\neq 1$ and $j\neq N$. $\widetilde{A}^1_{1,1}$ is an upper-triangular matrix implies that $A^1_{N,N}\neq 0$. If $A^1_{N,1}=0$, then the last row of $A^1$ contains only one non-zero element $A^1_{N,N}$. So by Observation 2, we are done. 

Otherwise, according to Observation~1, $\widetilde{A}^1 _{N,N}$ can be arranged into an upper-triangular matrix by permutation. Hence, after the corresponding permutations, we may convert $A^1$ into a new form, denoted by $A^2$ with the property that $A^2_{i,j}=0$ when $i>j$ and $i\neq N$. Moreover, permutations
that convert $A^1$ to $A^2$ never switch row $N$ (column N) of $A^1$ with any other rows (columns). So only one of $\{A^2_{N, j}|j\neq N\}$ is not zero. If $A^2_{N,1}= 0$, along with $A^2_{i,1} = 0$ for $N>i>1$, we have that the first column of $A^2$ contains exactly one non-zero element, $A^2_{1,1}$. So by Observation~2, we are done.


Otherwise, $A^2_{N,1}\neq 0$. According to Observation~3, $A^2_{N,1}A^2_{1,k}A^2_{k,N}=0$, for $k=2, \dots, N-1$. So we have that $A^2_{1,k}A^2_{k,N}=0$, for $k=2, \dots, N-1$. We will proceed by analyzing cases from $k=2$ to $k=N-1$. 

When $k=2$, if $A^2_{1,2}=0$, then column~2 of $A^2$ contains only one non-zero element $A^2_{2,2}$, and we are done by Observation~2. Otherwise, we may assume that $A^2_{1,2}\neq 0$ and $A^2_{2,N}=0$. 

When $k=3$, if $A^2_{3,N}\neq 0$, then $A^2_{1,3}=0$.
According to Observation~3, $A^2_{N,1}A^2_{1,2}A^2_{2,3}A^2_{3,N}=0$, and this implies that $A^2_{2,3}=0$. Hence, column~3 of $A^2$ contains only one non-zero element, $A^2_{3,3}$, and again we are done by Observation~2. Otherwise, we may assume that $A^2_{3,N}=0$, and one of $\{A^2_{1,3}, A^2_{2,3}\}$ is not zero.

When $k=k$, if $A^2_{4,N}\neq 0$, then $A^2_{1,4}=0$. Similarly, as in the case where $k=3$ (by Observation~3), $A^2_{N,1}A^2_{1,2}A^2_{2,4}A^2_{3,N}=0$, and this implies that $A^2_{2,4}=0$. One of $\{A^2_{1,3}, A^2_{2,3}\}$ is not zero $\Longrightarrow$ either $A^2_{N,1}A^2_{1,3}A^2_{3,4}A^2_{3,N}=0$ or $A^2_{N,1}A^2_{1,2}A^2_{2,3}A^2_{3,4}A^2_{3,N}=0$ $\Longrightarrow$ $A^2_{3,4}=0$. 
Hence, column~4 of $A^2$ contains only one non-zero element, $A^2_{4,4}$, and again we are done by Observation~2. Otherwise, we may assume that $A^2_{4,N}=0$, and at least one of $\{A^2_{1,4}, A^2_{2,4}, A^2_{3,4}\}$ is not zero.

Inductively, either one of column~$k$'s of $A^2$ contains only one non-zero element, or $A^2_{k,N}=0$ for all $k=2, \dots, N-1$. Note that the latter case implies that column $N$ of $A^2$ contains only one non-zero element, $A^2_{N,N}$, as $A^2_{N, 1}\neq 0 \Longrightarrow A^2_{1, N}=0$. Either way, the proof is then completed by Observation~2.



\end{proof}


\subsection{Details to Example~\ref{poly}}

To construct $\Likmat$, first notice that if maximum likelihood is achieved, $\Likmat_{1,1} = \Likmat_{1,2}$ under all settings of $\Delta$, $a$, and $q$. This is because a first- and second-order polynomial give the same fit to $\data_1$.

For $\Likmat_{2,1}$, by symmetry arguments we know that the maximum-likelihood fit of a first-order polynomial to $\data_2$ is a horizontal line ($f(x)=b$).
We can find this value of $b$ through a grid search. 
Given this $b$,
\[
\Likmat_{2,1} = N_q(a;b)^2 N_q(-a;b)^2 N_q(\Delta + a;b)N_q(\Delta - a;b),
\]
where
\[
N_q(z;b) = \frac{\sqrt{\beta}}{C_q} e_q(-\beta (x_i - \mu)^2).
\]
Here, $\beta = \frac{1}{5-3q}$ so that the variance is 1; $e_q(x)$ is the $q$-exponential function defined by $[1 + (1 - q)x]^{\frac{1}{1-q}}$ when $q \neq 1$, and $\exp(x)$ when $q = 1$. The normalizing constant $C_q$ is given by:
\[
C_q =
\begin{cases}
\frac{2\sqrt{\pi}\Gamma(\frac{1}{1-q})}{(3 - q)\sqrt{1-q}\Gamma(\frac{3-q}{2(1-q)})} & \text{for } -\infty < q < 1 \\

\sqrt{\pi} & \text{for } q = 1 \\

\frac{\sqrt{\pi}\Gamma(\frac{3-q}{2(q-1)}}{\sqrt{q-1}\Gamma\frac{1}{q-1}} & \text{for } 1 < q < 3.
\end{cases}
\]

For $\Likmat_{2,2}$, again by symmetry arguments we know that the maximum-likelihood fit of a second order polynomial to $\data_2$ is a parabola that passes through the middle of each of the three pairs of data points. Thus, $\Likmat_{2,2} = N_q(a;0)^6$.

\end{document}